\documentclass[twocolumn,english,journal,letterpaper,oneside,twocolumn]{IEEEtran}
\usepackage[T1]{fontenc}
\usepackage{babel}
\usepackage{array}
\usepackage{verbatim}
\usepackage{float}
\usepackage{multirow}
\usepackage{amsmath}
\usepackage{amssymb}
\usepackage{graphicx}
\usepackage{xcolor}
\usepackage{color}
\usepackage{gensymb}
 
\usepackage{caption}
\usepackage{times}
\usepackage{xcolor}
\usepackage{soul}
\usepackage[utf8]{inputenc}

\usepackage{epsfig}
\usepackage{graphicx}
\usepackage{amsmath}
\usepackage{amssymb}
\usepackage[utf8]{inputenc} 
\usepackage[T1]{fontenc}    
\usepackage{url}            
\usepackage{booktabs}       
\usepackage{amsfonts}       
\usepackage{nicefrac}       
\usepackage{microtype}      
\usepackage{enumitem}
\usepackage{dsfont}

\definecolor{dg}{rgb}{0,0.694,0.298}
\definecolor{purple}{rgb}{0.4,0.176,0.569}

\definecolor{royalblue}{RGB}{65,105,225}
\usepackage[colorlinks,
linkcolor=red,
citecolor=royalblue
]{hyperref}

\usepackage{xspace}
\makeatletter
\DeclareRobustCommand\onedot{\futurelet\@let@token\@onedot}
\def\@onedot{\ifx\@let@token.\else.\null\fi\xspace}
\def\eg{\emph{e.g}\onedot} 
\def\ie{\emph{i.e}\onedot} 
 
\def\etc{\emph{etc}\onedot}

\makeatother

\usepackage{pifont}
\usepackage{scalerel}
\usepackage{tikz}
\usetikzlibrary{svg.path}
\definecolor{orcidlogocol}{HTML}{A6CE39}
\tikzset{
  orcidlogo/.pic={
    \fill[orcidlogocol] svg{M256,128c0,70.7-57.3,128-128,128C57.3,256,0,198.7,0,128C0,57.3,57.3,0,128,0C198.7,0,256,57.3,256,128z};
    \fill[white] svg{M86.3,186.2H70.9V79.1h15.4v48.4V186.2z}
                 svg{M108.9,79.1h41.6c39.6,0,57,28.3,57,53.6c0,27.5-21.5,53.6-56.8,53.6h-41.8V79.1z M124.3,172.4h24.5c34.9,0,42.9-26.5,42.9-39.7c0-21.5-13.7-39.7-43.7-39.7h-23.7V172.4z}
                 svg{M88.7,56.8c0,5.5-4.5,10.1-10.1,10.1c-5.6,0-10.1-4.6-10.1-10.1c0-5.6,4.5-10.1,10.1-10.1C84.2,46.7,88.7,51.3,88.7,56.8z};
  }
}
\newcommand\orcidicon[1]{\href{https://orcid.org/#1}{\mbox{\scalerel*{
\begin{tikzpicture}[yscale=-1,transform shape]
\pic{orcidlogo};
\end{tikzpicture}
}{|}}}}

\makeatletter

\providecommand{\tabularnewline}{\\}
\floatstyle{ruled}
\newfloat{algorithm}{tbp}{loa}
\providecommand{\algorithmname}{Algorithm}
\floatname{algorithm}{\proV1tect\algorithmname}

\@ifundefined{date}{}{\date{}}
\makeatother

\begin{document}

\title{\emph{Let There be Light}: Improved Traffic Surveillance via Detail Preserving Night-to-Day Transfer}


\author{Lan Fu\,\orcidicon{0000-0003-3443-2381}\,,~\IEEEmembership{Student~Member,~IEEE,}
        Hongkai Yu\,\orcidicon{0000-0001-5383-8913}\,,~\IEEEmembership{Member,~IEEE,}
        Felix Juefei-Xu\,\orcidicon{0000-0002-0857-8611}\,,~\IEEEmembership{Member,~IEEE,}\\
        Jinlong Li\,\orcidicon{0000-0002-7784-8363}\,,  
        Qing Guo\,\orcidicon{0000-0003-0974-9299}\IEEEauthorrefmark{2},~\IEEEmembership{Member,~IEEE,} 
        and~Song~Wang\,\orcidicon{0000-0003-4152-5295}\IEEEauthorrefmark{2},~\IEEEmembership{Senior~Member,~IEEE}
\thanks{Lan Fu and Song Wang are with the Department of Computer Science and Engineering, University of South Carolina, SC, USA, e-mail: lanf@email.sc.edu, songwang@cec.sc.edu.} 
\thanks{Hongkai Yu and Jinlong Li are with the Department of Electrical Engineering and Computer Science, Cleveland State University, Cleveland, OH, USA, e-mail: h.yu19@csuohio.edu and lijinlong1117@foxmail.com.} 
\thanks{Felix Juefei-Xu is with the Alibaba Group, Sunnyvale, CA, USA, e-mail: juefei.xu@gmail.com.} 
\thanks{Qing Guo is with the Nanyang Technological University, Singapore, e-mail: tsingqguo@ieee.org.}%
\thanks{$\dagger$ indicates the co-corresponding authors: Qing Guo and Song Wang.} 
}

\maketitle

\begin{abstract}
In recent years, image and video surveillance have made considerable progresses to the Intelligent Transportation Systems (ITS) with the help of deep Convolutional Neural Networks (CNNs). As one of the state-of-the-art perception approaches, detecting the interested objects in each frame of video surveillance is widely desired by ITS. Currently, object detection shows remarkable efficiency and reliability in standard scenarios such as daytime scenes with favorable illumination conditions. However, in face of adverse conditions such as the nighttime, object detection loses its accuracy significantly. One of the main causes of the problem is the lack of sufficient annotated detection datasets of nighttime scenes. In this paper, we propose a framework to alleviate the accuracy decline when object detection is taken to adverse conditions by using image translation method. We propose to utilize style translation based StyleMix method to acquire pairs of day time image and nighttime image as training data for following nighttime to daytime image translation. To alleviate the detail corruptions caused by Generative Adversarial Networks (GANs), we propose to utilize Kernel Prediction Network (KPN) based method to refine the nighttime to daytime image translation. The KPN network is trained with object detection task together to adapt the trained daytime model to nighttime vehicle detection directly. Experiments on vehicle detection verified the accuracy and effectiveness of the proposed approach.
\end{abstract}

\begin{IEEEkeywords} nighttime object detection, image translation, kernel prediction network
\end{IEEEkeywords}

\section{Introduction}\label{sec:intro}
\IEEEPARstart{W}{ith} the fast development of computer vision and deep Convolutional Neural Networks (CNNs), visual data understanding in image and video has attracted a lot of attention \cite{zhang2020unsupervised,zhang2018spftn,yang2018segmentation,zhang2018poseflow,han2018reinforcement,yu2020weakly,li2014video}. For example, in the Intelligent Transportation Systems (ITS), detecting the vehicles in each frame of the traffic surveillance video is important to extract the real-time traffic flow parameters \cite{ke2018real} for the efficient traffic control and obtain the vehicle trajectories \cite{chen2020high} for the calibrated traffic model simulation, \etc. 
\begin{figure*}[th!]
\centering
\includegraphics[width=0.8\textwidth]{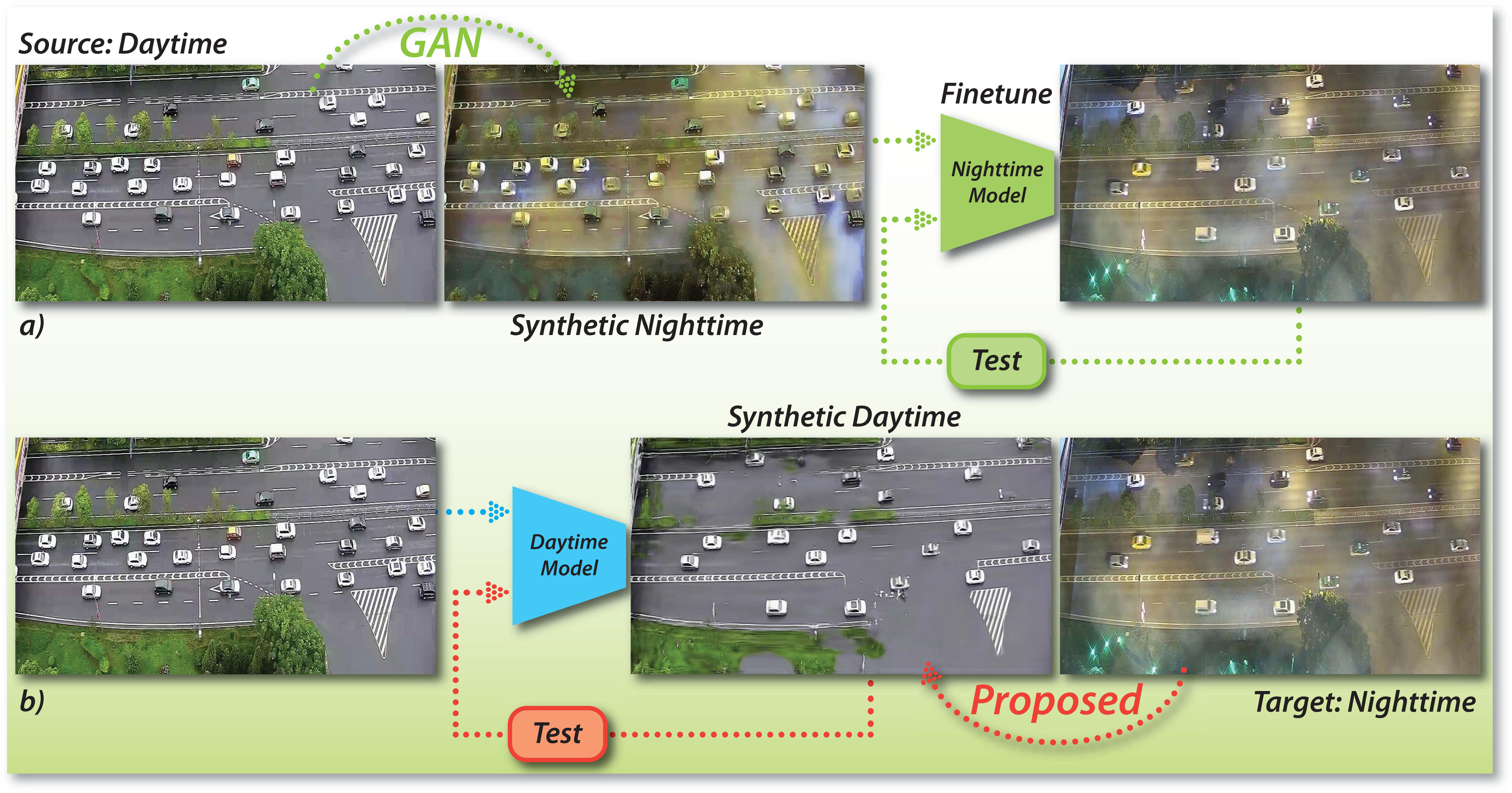}
\caption{Illustration of the domain reuse problem: a) traditional method with style transfer and the nighttime model fine-tuned from the daytime model, b) the proposed detail-preserving Night-to-Day translation method without changing the daytime model.}
\label{pipeline}
\end{figure*}

Most of existing researches focus on daytime perception task through supervised learning, however, they generalize badly on adverse conditions such as nighttime scenarios \cite{jinlong2020domain}. The adversity of nighttime scenario poses two challenges for the success of perception task at nighttime: 1) nighttime data with a large amount of annotations is usually scarce compared to the large-scale daytime data, since accurate annotation of nighttime images is relatively hard to obtain. 2) The visual hazards, such as underexposure and noise, of nighttime images cause the extracted features corrupted.

One traditional way to solve this is to fine-tune the already-trained daytime perception model on the limited nighttime data, and hopefully it can perform well on nighttime scenarios, but it requires extra time and additional-labeled nighttime data for model fine-tuning. Another traditional way \cite{jinlong2020domain} might use Generative Adversarial Networks (GANs) based image to image translation methods in an unpaired way, such as CycleGAN \cite{zhu2017unpaired} and UNIT \cite{liu2017unsupervised}, to transfer daytime images to fake nighttime images. Paired daytime and nighttime images are hard to obtain in the real-world applications, due to the dynamic traffic and environment changes. This kind of image translation considers this problem as domain adaptation for model fine-tuning on synthetic nighttime images without labeling the nighttime data. However, these methods also need extra time for model fine-tuning. 
In addition, GAN based image translation suffers from model collapse and does not preserve content details very well \cite{park2020contrastive, liu2017unsupervised, zhu2017unpaired, huang2018multimodal, wang2019learning}. Bottleneck layers in a general deep generator hurt the learning ability of convolution kernels due to downsampling and upsampling operations, resulting in possible losing some structure details. Besides, unpaired training data of different domains limits the detail-preserving ability of generators due to the lack of pixel-wise correspondence.  

In this paper, we would like to reuse the daytime perception model to nighttime scenarios. Our basic idea is to maximally use the pretrained daytime perception model, similar to works \cite{yao2016semantic,han2019p}, which could be easily extended to the nighttime tasks. Reversely to the traditional methods, we transfer the nighttime images back to the daytime style with the detail-preserving to reuse the trained daytime perception model, as shown in Figure~\ref{pipeline}. The strengths of this reverse way are obvious and promising: 1) there are no extra training efforts for the already trained daytime perception model and no needs to manually label the nighttime data; 2) image transfer could reduce the domain distribution discrepancy between daytime and nighttime data; 3) detail-preserving image transfer could better maintain the structure details than the GAN based image transfer. 

Specifically, we propose a detail-preserving unpaired domain transfer method for this task, which mainly contains two components: 1) Style-transfer based StyleMix, 2) Kernel Prediction Network (KPN) based nighttime to daytime image transfer. Without paired daytime-nighttime image pairs, we propose to utilize style translation based StyleMix method, inspired by AugMix \cite{hendrycks2019augmix}, to acquire pairs of daytime and nighttime images as training data for the following nighttime to daytime image transfer. We can effectively alleviate the detail corruption caused by GAN: 1) The synthetic nighttime image and corresponding daytime image translation can provide pixel-wise correspondence for night-to-day translation. 2) Kernel prediction network based method can refine the nighttime to daytime image translation because the per-pixel kernel fusion can effectively utilize the neighboring region for each pixel and could learn more spatial context representing structure information. The proposed method can conduct daytime and nighttime vehicle detection with just one daytime model, which is more convenient in real-world applications.

In this paper, we choose the vehicle detection problem in the traffic surveillance video as a  case study for the proposed approach. The KPN network is trained with object detection task together to adapt the trained daytime model to fit nighttime domain directly. Experimental results on a vehicle video dataset in daytime and nighttime verified the accuracy and effectiveness of the proposed approach. The contributions of this paper are summarized in the following: 

\begin{itemize}[noitemsep, nolistsep,leftmargin=*] 
    \item We propose a detail-preserving unpaired domain transfer method for nighttime vehicle detection to adapt the trained daytime model directly for nighttime vehicle detection.
    
    \item To solve the problem of lack of paired daytime-nighttime image pairs, we propose to utilize style translation based StyleMix method to obtain pairs of daytime image and nighttime image as training data. These training data are utilized by KPN network to perform nighttime to daytime image transfer.
    
    \item The comprehensive experimental results on a vehicle detection dataset from the video surveillance scenario in daytime and nighttime show that the proposed method achieved better vehicle detection performance in nighttime scenario.   
\end{itemize}

In the following of this paper, Section~\ref{sec:related} reviews the related work. Section~\ref{sec:method} explains the proposed method. Experiment setting and results are described in Section~\ref{sec:exp}, followed by a conclusion in Section~\ref{sec:concl}.

\section{Related Work}\label{sec:related}
\textbf{Object detection at nighttime:} The state-of-the-art performance in object detection is rapidly improving in recent years. One-stage (SSD \cite{liu2016ssd}, YOLO \cite{redmon2016you}, RetinaNet \cite{lin2017focal}) and two-stage (Faster R-CNN \cite{ren2015faster}, Mask R-CNN \cite{he2017mask}) detection frameworks have achieved promising performance in real-world applications. They generally require a large amount of manually labeled data for supervised learning. Nonetheless, most of them operate well at daytime, under favorable illumination conditions, and scale badly to nighttime scenarios with challenging lighting conditions. Further, manual annotation of nighttime images are hard and time-consuming, because even human cannot clearly discern objects in adverse nighttime scenario. Nighttime detection task has attracted a lot of attention recently. Domain specific works \cite{xu2005pedestrian,ge2009real,choi2018kaist} explore the human detection at nighttime by considering the type of cameras. Other works \cite{kuang2017bayes,satzoda2016looking} pertain to vehicle detection in driving scenarios. Domain-invariant representations \cite{alvarez2010road,ros2015unsupervised} or fusion works \cite{valada2017adapnet} are designed to be robust to illumination changes. Image translation work \cite{anoosheh2019night} aims to improve retrieval-based localization at nighttime. In this work, we focus on vehicle detection at nighttime in traffic surveillance scenarios. We aim to adapt the daytime detection model to nighttime detection for reusing the daytime domain knowledge. 
It is also interesting to explore the robustness of daytime model from other perception tasks \cite{tian2020bias,neurips20_abba}, \eg, object tracking \cite{guo2021exploring,guo2020selective,guo2020spark,guo2017learning}, at nighttime scenario in the future.

\textbf{Learning from synthetic data:} In general, CNNs perform much worse when there is a domain shift between training and testing sets, which hurts the generalization ability of CNNs. Data augmentation techniques, like random cropping and affine transformation, are one way to improve the stability of networks in unfamiliar domains. Effective use of synthetic  data \cite{wang2019learning,zhang2018fully,bak2018domain} is another choice to achieve the same goal. It has been used in a lot of computer vision tasks such as crowd  counting \cite{wang2019learning}, semantic  segmentation \cite{zhang2018fully}, person re-identification \cite{bak2018domain}, \etc. One common way of reducing dataset distribution bias is to make the synthetic data much more photo-realistic and minimize the domain shift at the same time. Some domain adaptation methods \cite{zhang2020cross,kim2020learning,zheng2020cross} can be used to learn the domain-invariant features to align the domains of synthetic and real data, so the model generalization ability can be improved. In this paper, we utilize pairs of synthetic nighttime and real daytime images based on style-transfer method for training a detail-preserving night-to-day network. Then, we adapt our trained translation network to transfer any nighttime image  to daytime version so as to reuse the trained daytime detection  model.


\textbf{Style transfer:} Many computer vision tasks need to  translate an input image from one domain to another domain, which are viewed as the image translation problem. Generative adversarial networks (GANs) based methods are promising for image stylization, which aim to sample from a probability distribution to generate images. GANs include two models: a generative model  and a discriminative model. The former captures the critical data distribution for image generation, while the latter aims to distinguish between real and generated samples.  CycleGAN \cite{zhu2017unpaired} extends the GAN-based image translation method to an unsupervised framework, where no paired data is required. It performs a full translation cycle both from  source domain to target domain and from target domain back to source domain, and then it could regularize the high cycle  consistency. Subsequent works encourage a shared latent feature space via a variational autoencoder (UNIT \cite{liu2017unsupervised}). ComboGAN \cite{anoosheh2018combogan} and SMIT \cite{romero2019smit} extend to multiple domain translation. GcGAN \cite{fu2019geometry} proposed that the translation network should keep geometry consistency.

Although unpaired image translation by GAN-based methods are popular for style transfer, the generated images might lack details due to the common existing downsampling and upsampling network operations. In this paper, we propose to train a detail-persevering network to achieve the unpaired domain transfer for nighttime object detection. 

\section{Methodology}\label{sec:method}

In this section, we propose the \textit{detail-preserving unpaired domain transfer} for performing high accuracy object detection in the nighttime without retraining the detectors on daytime dataset. We introduce the whole framework in Sec.~\ref{subsec:method_overview} and reveal the challenges. Then, our two main contributions, \ie, \textit{scene-aware pixel-wise filtering} in Sec.~\ref{subsec:kpn} and \textit{StyleMix} in Sec.~\ref{subsec:stylemix}, help 
to address the challenges and achieve much better detection accuracy.

\subsection{Detail-preserving Unpaired Domain Transfer for Nighttime Object Detection}\label{subsec:method_overview}

\begin{figure}[th!]
\centering
\includegraphics[width=1.0\columnwidth]{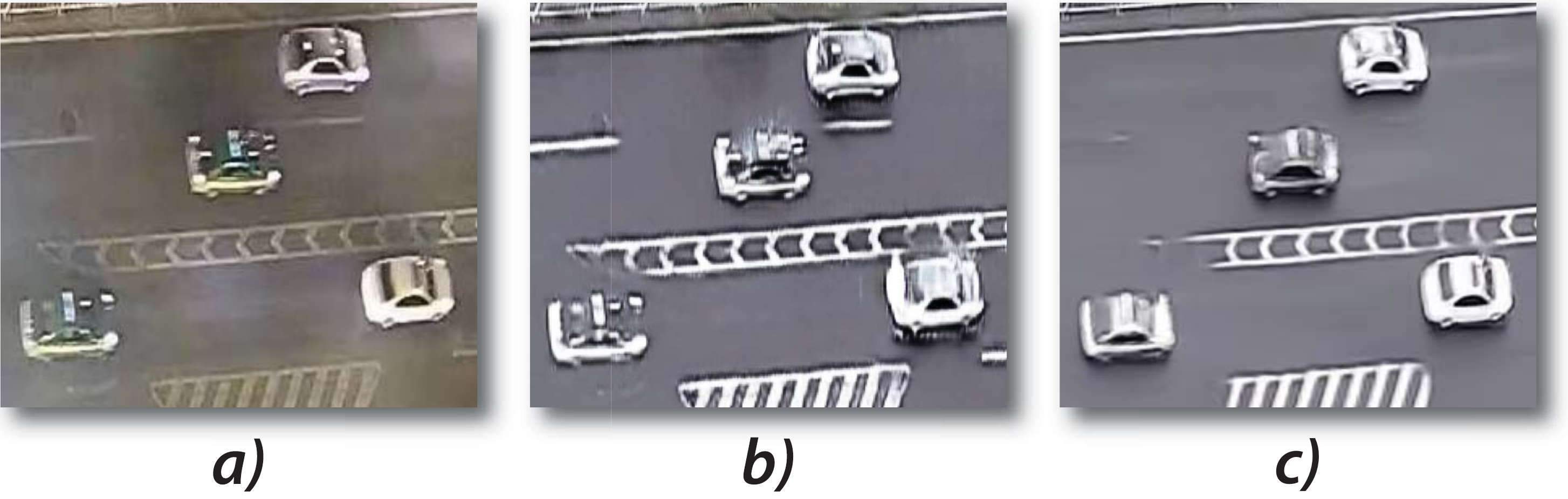}
\caption{Image translation results of GAN-based method and the proposed method: a) target nighttime image, b) translated daytime image of a) by GAN-based method \textit{CycleGAN}~\cite{zhu2017unpaired}, c) translated daytime image of a) by the proposed method.}
\label{fig:detail}
\end{figure}

We propose to perform nighttime object detection by transferring input nighttime images to the corresponding daytime versions for further object detection. This task could be simply formulated as
%
\begin{align}\label{eq:n2d}
\hat{\mathbf{I}}=\phi(\mathbf{I}),
\end{align}
%
where the $\phi(\cdot)$ denotes a transfer function that can map the nighttime image $\mathbf{I}$ to the corresponding daytime version. 
A straightforward way is to set $\phi(\cdot)$ as a popular generator that can be trained with the adversarial loss. Nevertheless, we argue that GAN-based transfer is hard to recover the details in the nighttime image, which is rather important for accurate object detection. As shown in Figure~\ref{fig:detail}, the GAN-based method might destroy the detailed car structure, leading to missing detection. Actually, the night-to-day translation for object detection requires that the object-related details, \eg, car's structure, should be preserved while different scene patterns in the night should be perceived and properly mapped to its daytime versions, posing two challenges for deep learning-based solutions: \ding{182} Popular deep generator based methods easily harm the object details due to the common existing bottleneck layers where the input image is transferred by downsampling and upsampling. \ding{183} It is hard to get paired dataset which is significantly important for training detail-persevering networks with pixel-wise correspondence.

\begin{figure*}[th!]
\centering
\includegraphics[width=0.9\textwidth]{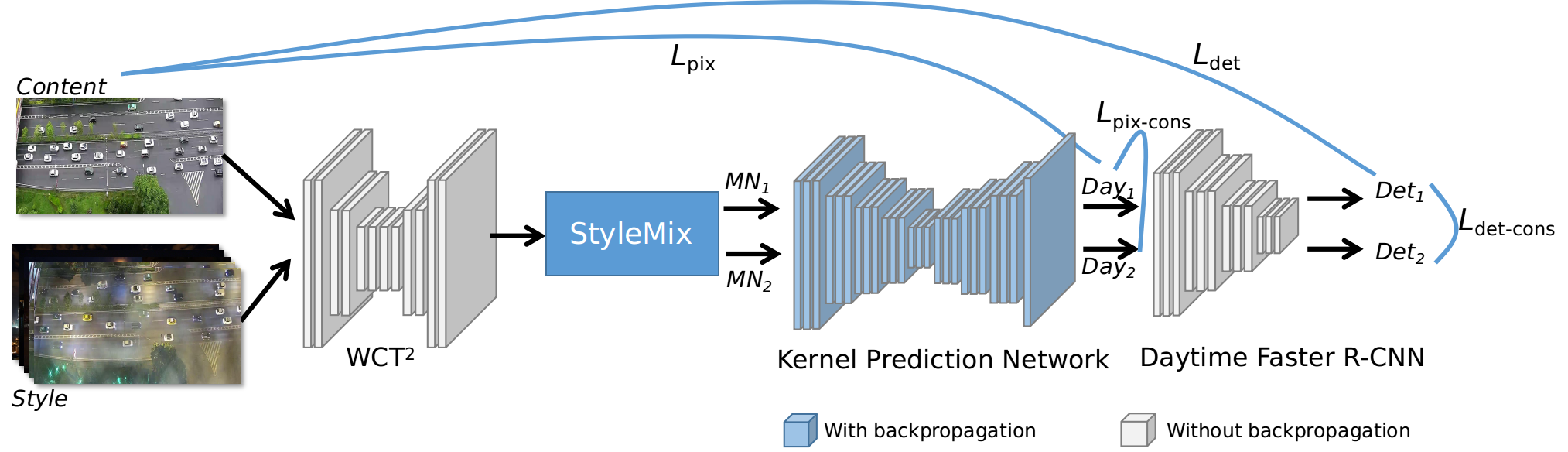}
\caption{The proposed object detection pipeline at night with Night-to-Day image translation.}
\label{fig:pipeline}
\end{figure*}
To address the first issue, we propose the \textit{scene-aware pixel-wise filtering} in Sec.~\ref{subsec:kpn} for night-to-day transformation. Different to all existing works that employ a DNN as the transformer directly, our method maps the input image through a single-layer filtering whose kernels are predicted by an offline-trained DNN denoted as the \textit{kernel prediction network}. Note that, the single-layer filtering (without any downsampling and upsampling operations) avoids the risk of missing important object-related details. Meanwhile, the DNN helps understand the scene and predict spatial-variant kernels for effective transformation. Specifically, kernel prediction network predicts a kernel for each pixel to capture local spatial context information to preserve more details, \eg, structure information. Recent works  \cite{cai2019toward,mildenhall2018burst, guo2020efficientderain, fu2021auto} have proved that per-pixel kernel prediction network can achieve image recovery with better details. 
To address the second challenge, we propose a style-transfer-based data augmentation method, \ie, \textit{StyleMix} in Sec.~\ref{subsec:stylemix}, to generate nighttime-daytime image pairs for training the kernel prediction network.

We show the whole framework in Figure~\ref{fig:pipeline}. Intuitively, our method is a pre-process module transferring the input nighttime image to daytime version for further object detection, which is supported by a novel and simple data augmentation method, \ie, StyleMix.

\subsection{Scene-aware Pixel-wise Filtering} \label{subsec:kpn}

We propose the scene-aware pixel-wise filtering for the night-to-day transformation. Specifically, we reformulate Eq.~\eqref{eq:n2d} as
%
\begin{align}\label{eq:kpn}
\hat{\mathbf{I}} = &\mathbf{K}\circledast\mathbf{I}, \\
\mathrm{with~~} &\mathbf{K} = \phi(\mathbf{I}),
\end{align}
%
where $\circledast$ denotes the pixel-wise filtering, $\mathbf{K}$ is a pixel-wise filter $\in$ $\mathds{R}^{(k\times k)\times h\times w}$. Each vector in channel dimension $\mathbf{K}(i,j)$ $\in$ $\mathds{R}^{(k\times k)}$ is a per-pixel kernel and can be applied to the $k \times k$ neighborhood region of each pixel in the input nighttime image $\mathbf{I}$ by element-wise multiplication. The $\phi(\cdot)$ denotes the kernel prediction network and is used to 
perceive the input image and predict the suitable kernel for each pixel. 

Then, we acquire the daytime version $\hat{\mathbf{I}}$ of the input image. Since it is pixel-wise filtering of the input nighttime image, it could largely preserve the image details without corruption. To fully leverage rich neighborhood information of every image pixel, a large kernel size $k$ is desired, however, the computational and memory cost will increase as well. The kernel size $k$ in our implementation is set to 5. 

\begin{figure}[ht]
\centering
\includegraphics[width=1.0\columnwidth]{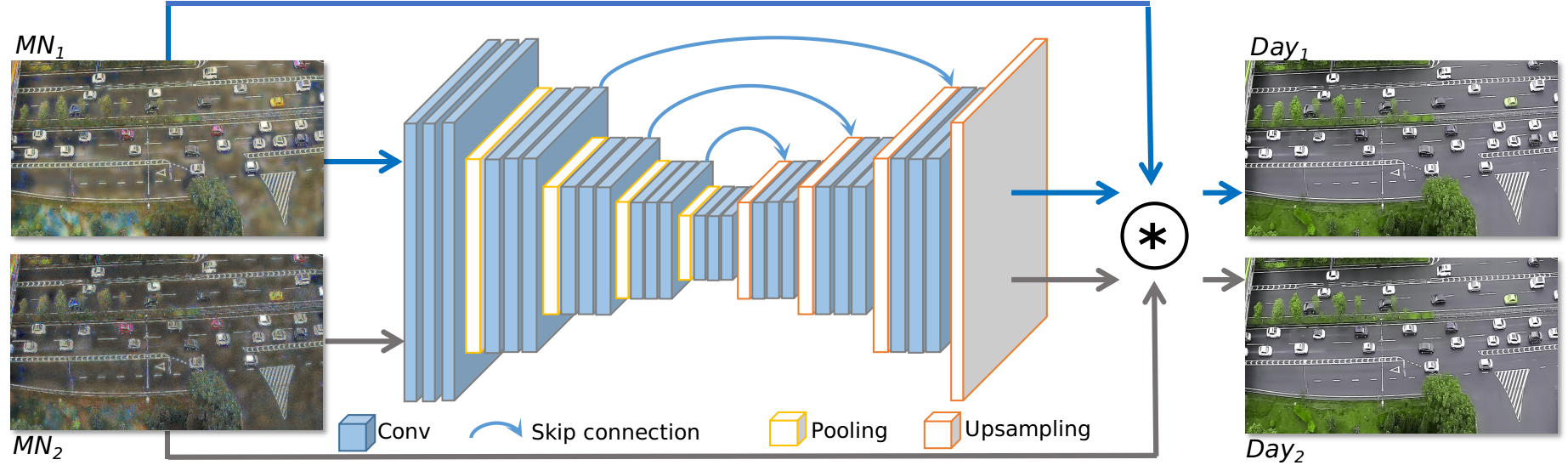}
\caption{Illustration of the kernel prediction network based scene-aware pixel-wise filtering.}
\label{fig:kpn}
\end{figure}

The framework of kernel prediction network is shown in Figure~\ref{fig:kpn}. In this work, the training input data for KPN is synthetic nighttime images from Sec.~\ref{subsec:stylemix}. Specifically, two synthetic Mixed Nighttime images $\text{MN}_{1}$ and $\text{MN}_{2}$ with different style conditions are fed into KPN, respectively. KPN will output image-specific per-pixel filter for each image, respectively. Then element-wise multiplying the specific filter with the corresponding input image will generate the daytime version image $\hat{\mathbf{I}}_{i}$, $i$ = $\text{1, 2}$.

The basic loss function $\mathcal{L}_\mathrm{pix}(\hat{\mathbf{I}}_{i}, \hat{\mathbf{I}}^*)$ is the pixel-wise $L_1$ distance between the ground truth daytime image $\hat{\mathbf{I}}^*$ and the translated daytime image $\hat{\mathbf{I}}_{i}$. It is defined as
%
\begin{align}\label{eq:l1}
\mathcal{L}_\mathrm{pix}(\hat{\mathbf{I}}_{i}, \hat{\mathbf{I}}^*) =\|\hat{\mathbf{I}}^*-\hat{\mathbf{I}}_{i}\|_{1}.
\end{align}
%
We also define a consistency loss $\mathcal{L}_\mathrm{pix-cons}$ between $\hat{\mathbf{I}}_{1}$ and $\hat{\mathbf{I}}_{2}$ by measuring their $L_1$ distance. The equation is 
%
\begin{align}\label{eq:cons}
\mathcal{L}_\mathrm{pix-cons}(\hat{\mathbf{I}}_{1}, \hat{\mathbf{I}}_{2}) =\|\hat{\mathbf{I}}_{1}-\hat{\mathbf{I}}_{2}\|_{1}.
\end{align}
%

\subsection{StyleMix: Bridging the Gap to Nighttime Data}\label{subsec:stylemix}
The style-transfer-based method is utilized to generate nighttime-daytime image pairs for KPN training. To bridge the shift of synthetic nighttime and target nighttime data, we propose the SytleMix strategy to embody the diversity of nighttime scenarios. 

\begin{figure*}[t]
\centering
\includegraphics[width=0.8\textwidth]{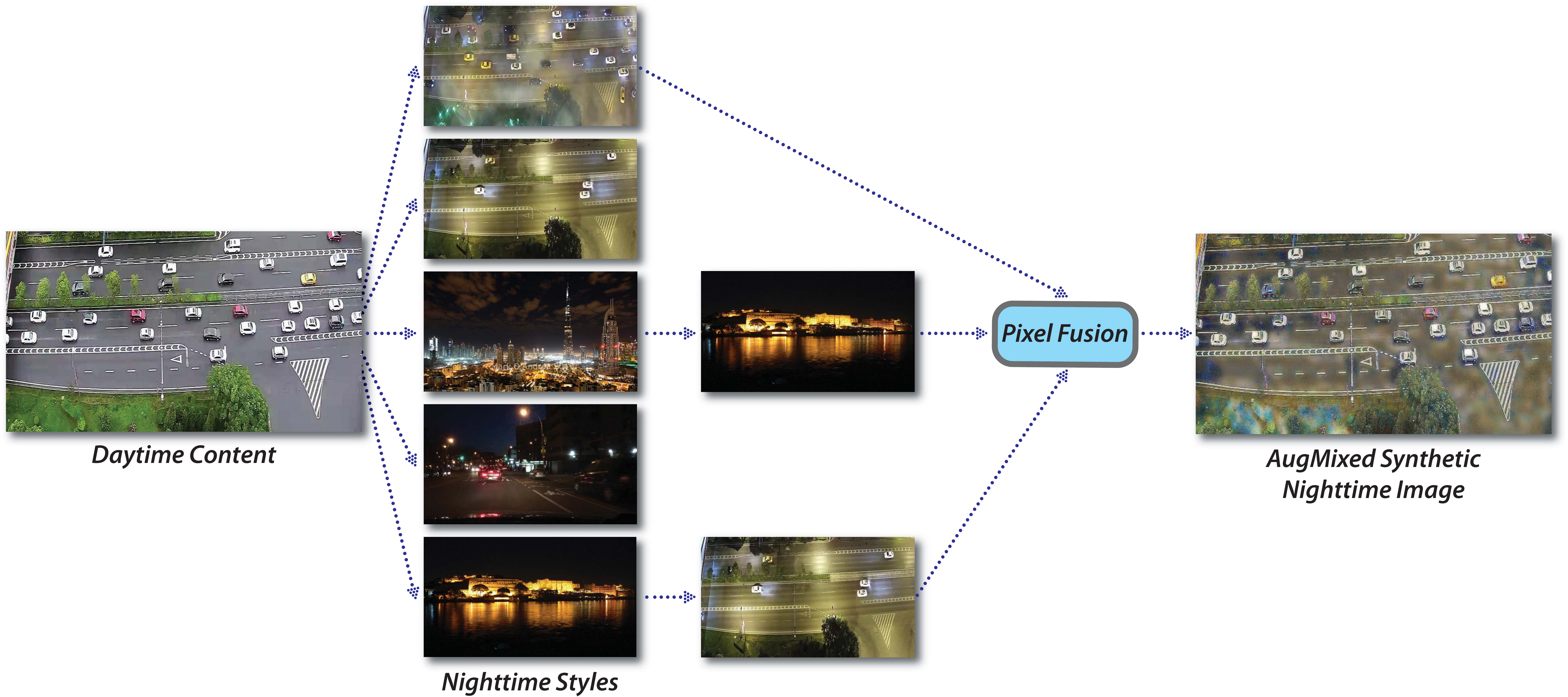}
\caption{Illustration of the proposed StyleMix method to bridge the gap to nighttime data.}
\label{fig:stylemix}
\end{figure*}

\begin{figure}[ht]
\centering
\includegraphics[width=1.0\columnwidth]{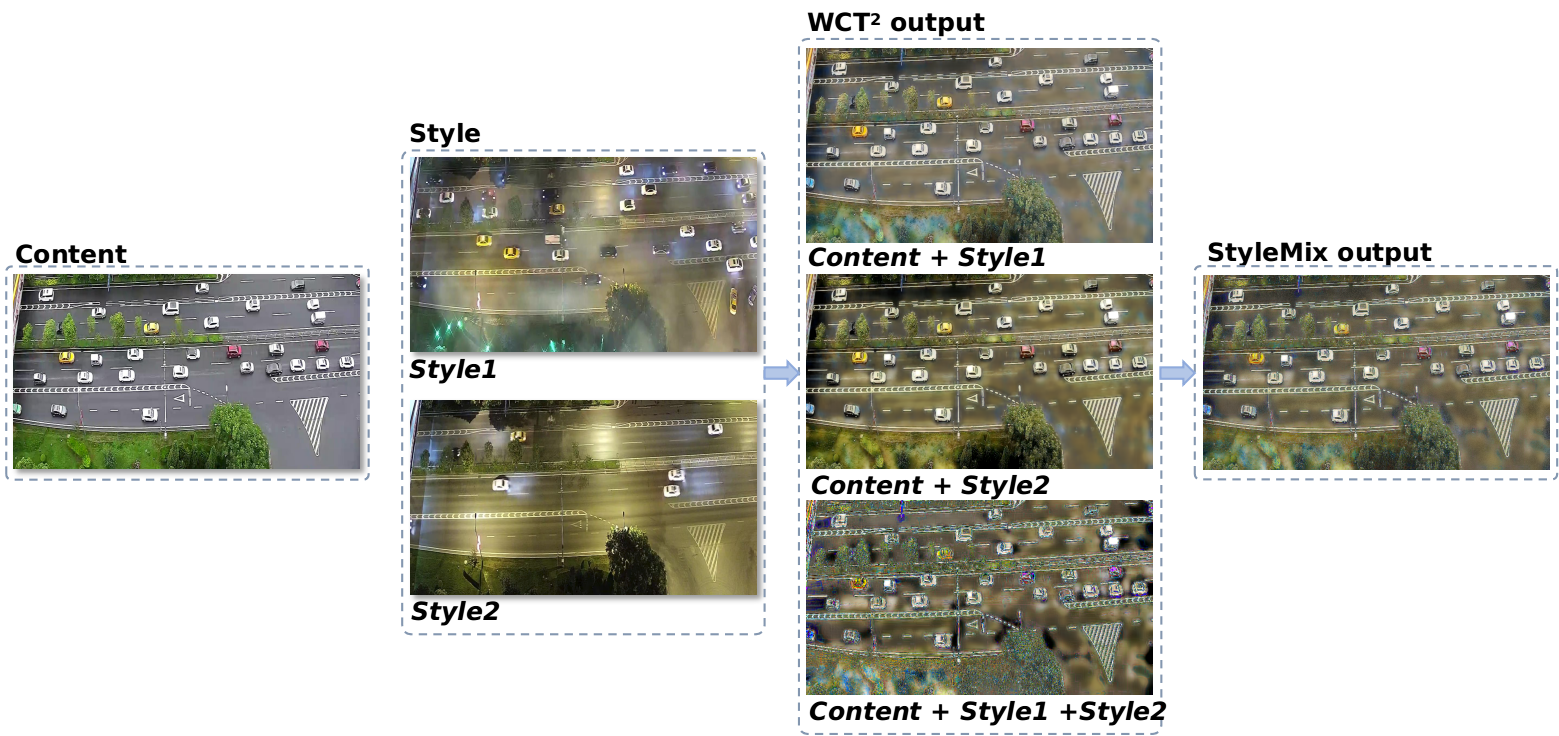}
\caption{Sample visualization of the proposed StyleMix to generate synthetic nighttime images from real daytime images.}
\label{fig:mixs}
\end{figure}

Specifically, style-transfer network can preserve the structure of input content image and stylize the content image according to the input style reference to implement image translation. As shown in Figure~\ref{fig:pipeline}, we adopt a pretrained style transfer network, whitening and coloring transforms ($\text{WCT}^2$), to finish daytime to nighttime image translation. For the input of $\text{WCT}^2$, daytime images are the content image and five real nighttime images act as the style reference. Five style reference images for following StyleMix are selected depending on the illumination condition of target nighttime scenarios. During daytime to nighttime image translation, StyleMix is involved to reduce the distribution shift of translated style and target nighttime style. It works in the way shown in Figure~\ref{fig:stylemix}. Specifically, for each daytime input image, three style augmentation chains out of fiver style reference are randomly sampled, each of which consists of one to two randomly selected style transfer operations. Then the transferred images from these style augmentation chains are combined by pixel-level fusion to acquire mixed nighttime image $\text{MN}_{i}$ in Sec.~\ref{subsec:kpn}. Pixel-wise fusion is implemented by pixel-wise convex operations between translated nighttime images and convex coefficients. We randomly sample from a Dirichlet $(\alpha, \ldots, \alpha)$ distribution to construct the 3-dimensional vector of pixel-wise convex coefficients. Figure~\ref{fig:mixs} shows one example of the pixel-wise fused synthetic nighttime image by StyleMix. The augmixed output of StyleMix are the pixel-wise fusion result of translated content image with different styles. It shows that StyleMix can effectively generate  various kinds of synthetic nighttime images which are visually close to the real nighttime scene.

For detection task of the pipeline, we construct the detection loss $\mathcal{L}_\mathrm{det}({\textit{Det}}_{i},  {\textit{Det}}^*)$ and the detection consistency loss  $\mathcal{L}_\mathrm{det-cons}({\textit{Det}}_{1}, {\textit{Det}}_{2})$. We adopt Smooth $L_1$ loss~\cite{girshick2014rich} to calculate $\mathcal{L}_\mathrm{det}$ and $\mathcal{L}_\mathrm{det-cons}$. The total loss $\mathcal{L}_\mathrm{N2D}$ of the pipeline is a weighted sum of $\mathcal{L}_\mathrm{pix}$, $\mathcal{L}_\mathrm{pix-cons}$, $\mathcal{L}_\mathrm{det}$, and $\mathcal{L}_\mathrm{det-cons}$. It is defined as
%
\begin{align}
\mathcal{L}_\mathrm{N2D} = \mathcal{L}_\mathrm{pix} + \mathcal{L}_\mathrm{pix-cons} + \lambda( \mathcal{L}_\mathrm{det} + \mathcal{L}_\mathrm{det-cons} ),
\end{align}
%
where $\lambda$ is set to 10 in our experiments.

\section{Experiments}\label{sec:exp}
\subsection{Datasets}
In this paper, the public D\&N-Car Benchmark~\cite{jinlong2020domain} is utilized to verify the effectiveness of the proposed approach. It is a real traffic surveillance dataset in urban expressway scene recorded in the city Xi'an, China. This dataset includes 1,200 daytime images and 1,000 nighttime images with their ground truth in the format of bounding boxes across different periods and dates, each of which is with resolution 1,280$\times$720. There are total 57,059 vehicle instances in this dataset. The training set consists of 1,000 daytime traffic images with manual ground-truth labels, denoted as \textbf{Day-training}. The testing set includes 1,200 images, where 200 images are in daytime and 1,000 images are in nighttime. In the 200 daytime testing images, 100 images are in the normal traffic condition, denoted as \textbf{Day-normal}, and the other 100 images are in the congested traffic condition, denoted as \textbf{Day-congested}. The left 1,000 images of testing set constitute 4 subsets of nighttime traffic images (denoted as \textbf{Night1}, \textbf{Night2}, \textbf{Night3}, \textbf{Night4}). The details of the benchmark are shown in Table~\ref{tab:data}. In the experiment, we denote the labeled daytime traffic images (Day-training) as the Source Domain \textbf{S}, and the unlabeled nighttime traffic images as the Target Domain \textbf{T}. 

\begin{table}[htbp]
\centering
\footnotesize
\caption{Details of the D\&N-Car Benchmark~\cite{jinlong2020domain}.}\label{tab:data}
\resizebox{0.9\linewidth}{!}{
\begin{tabular}{l|ccc}
\toprule
~  & $\text{No. of images}$ & $\text{No. of car instances}$ & \text{Time} \tabularnewline
\midrule
\text{Day-training} & 1000  & 32,456 & 19:10 \tabularnewline
\midrule
\text{Day-normal} & 100  & 3,173 & 19:00 \tabularnewline
\text{Day-congested} & 100  & 4,539 & 14:30 \tabularnewline
\midrule
\text{Night1} & 250  & 7,322 & 21:30 \tabularnewline
\text{Night2} & 250  & 5,554 & 21:30 \tabularnewline
\text{Night3} & 250  & 1,738 & 23:50 \tabularnewline
\text{Night4} & 250  & 2,277 & 00:20 \tabularnewline
\bottomrule
\end{tabular} 
}
\end{table}

\subsection{Experimental Setting}
We conduct experiments on two different scenarios: 1). Detect the vehicles during daytime by Faster R-CNN \cite{ren2015faster} model trained on \textbf{Day-training}; 2). Detect the vehicles during nighttime by trained Faster R-CNN model on  \textbf{Day-training} after the proposed  night-to-day image translation. The detailed experimental setting is as follows: 1) Scenario 1: We directly train a Faster R-CNN model on the dataset  \textbf{Day-training} in a supervised way and test the images on \textbf{Day-normal} and \textbf{Day-congested}, respectively; 2) Scenario 2: For style-transfer-based StyleMix image translation aiming at acquiring pairs of daytime and nighttime images, we utilize 1,000 images in \textbf{Day-training} set and 5 style reference images for nighttime images synthesis and augmixing styles. For KPN-based night-to-day training, there are 2,000 augmixed nighttime images for training in each epoch. Next, predicted daytime images are fed into detection task to further fit the translated daytime image for object detection. For inference, the trained KPN operates image translation for real nighttime images (\textbf{Night1}, \textbf{Night2}, \textbf{Night3}, and \textbf{Night4}), and then trained daytime detection model tests on translated nighttime images for performance evaluation. 

We set the method that directly tests on nighttime images with trained daytime model Faster R-CNN \cite{ren2015faster} as a baseline. We also compare the proposed method with unpaired image translation methods UNIT \cite{liu2017unsupervised}, CycleGAN \cite{zhu2017unpaired}, and GcGAN \cite{fu2019geometry} combining with Faster R-CNN in both day-to-night and night-to-day directions. To train the image translation models, the training dataset for daytime is the  \textbf{Day-training} set and the training set for nighttime is a combination of \textbf{Night1}, \textbf{Night2}, \textbf{Night3}, and \textbf{Night4}.

\begin{figure*}[th!]
\centering
\includegraphics[width=1.0\textwidth]{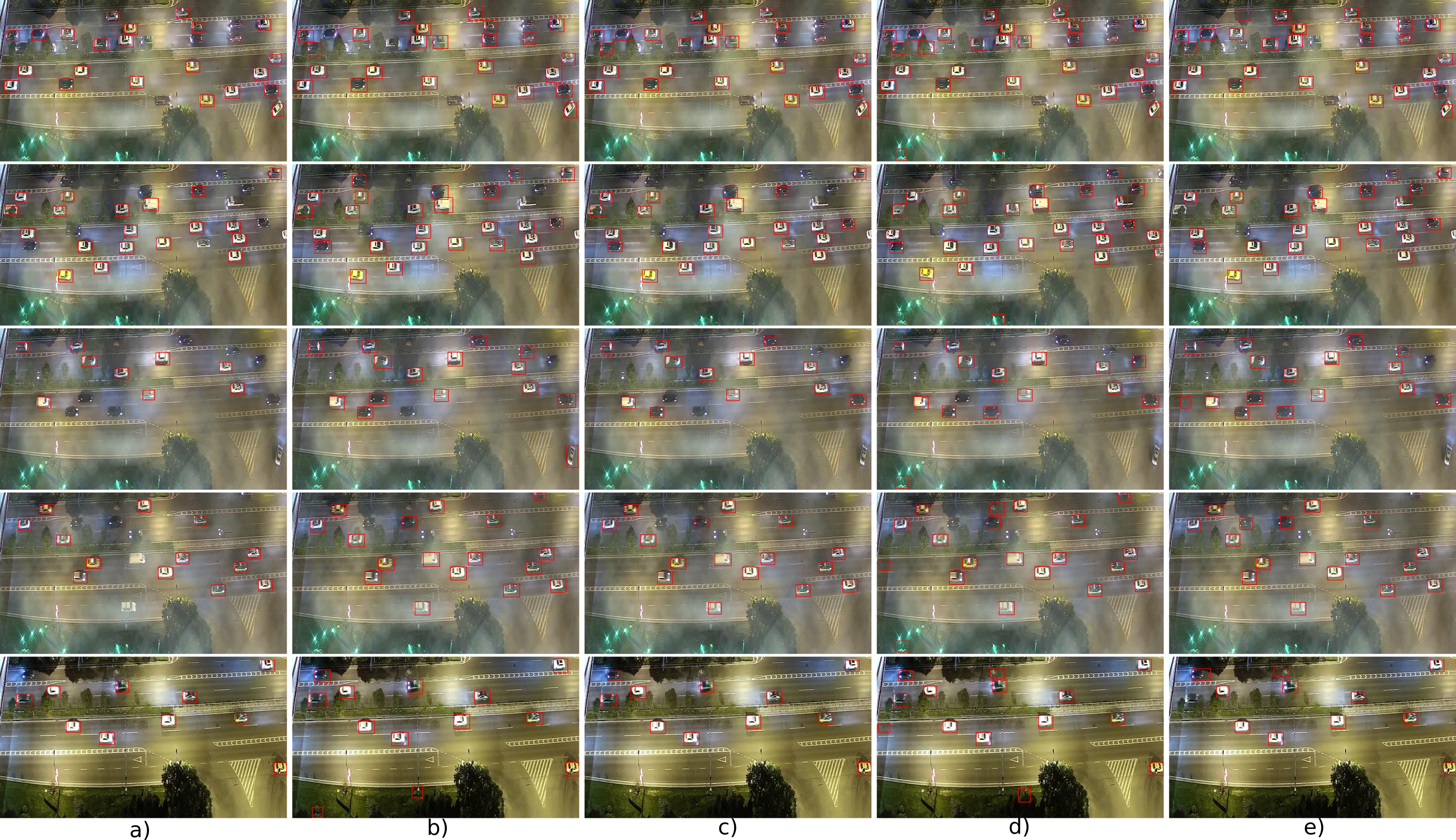}
\caption{Visualization results of nighttime vehicle detection. a)-e) are the detection results from Faster R-CNN \cite{ren2015faster}, ${\text{Faster R-CNN}_{\textit{n}}}$ + $\text{UNIT}_{d2n}$ \cite{liu2017unsupervised}, ${\text{Faster R-CNN}_{\textit{n}}}$ + $\text{CycleGAN}_{d2n}$ \cite{zhu2017unpaired}, ${\text{Faster R-CNN}_{\textit{n}}}$ + $\text{GcGAN}_{d2n}$ \cite{fu2019geometry}, and the proposed method, respectively. Note: red bounding box indicates detection result.}
\label{fig:vis_n}
\end{figure*}

We built our translation and detection pipeline in PyTorch. For object detection, we use ResNet50 as our backbone. For detection training, we utilize Stochastic Gradient Descent (SGD) to optimize our network and set the initial learning rate to 0.0001 and decay it after every 10 epochs. The experiments are conducted on a NVIDIA GTX 1080Ti GPU. For night-to-day image translation training, we train KPN with SGD by setting the learning rate to 0.002 for 200 epochs on two Tesla V100 GPUs. For a comprehensive performance evaluation, the widely-used object detection metric mAP (mean average precision) is used for evaluating the vehicle detection results. For all the experiments, the performance evaluation uses a uniform threshold of 0.5 for the Intersection Over Union (IoU) between the predicted bounding boxes and ground truth.

\subsection{Results on Benchmark}\label{sec:daytime}
We first report the detection results of one-stage detector SSD \cite{liu2016ssd} and two-stage detector Faster R-CNN \cite{ren2015faster} for Scenario 1, shown in Table~\ref{tab:eval_daytime}. We can see that both of the mAP drop from about 99\% to 88\% when the traffic is congested. The congested situation increases the object detection difficulty, resulting in a lower detection performance compared to the uncrowded situation. Because there is not a clear difference in terms of mAP between SSD and Faster R-CNN, we choose Faster R-CNN as our baseline detector for following experiments.

\begin{table}[htbp]
\centering
\caption{Daytime vehicle detection results.}\label{tab:eval_daytime}
\resizebox{0.65\linewidth}{!}{
\begin{tabular}{l|cccccc}
\toprule
$\text{mAP(\%)}$ 
& $\text{SSD}$  & $\text{Faster R-CNN}$\tabularnewline
\midrule
$\text{Day-normal}$ 
& 99.05 & 99.01 \tabularnewline
$\text{Day-congested}$ 
& 88.35 & 88.57 \tabularnewline
\bottomrule
\end{tabular}
}
\end{table}

\subsection{Results compared to day-to-night translation methods}
We compare the detection results of nighttime vehicle to other image translation methods in a day-to-night direction. According to Scenario 2, the proposed method performs vehicle detection on translated daytime images obtained from KPN by the daytime model obtained from Sec. \ref{sec:daytime}. However, for comparison methods performing nighttime vehicle detection in a day-to-night direction, they require additionally training a nighttime model for nighttime vehicle detection, besides the daytime model. For example, taking CycleGAN as the day-to-night image translation method, we translate daytime images to fake/synthetic nighttime images in an unpaired way, and followed by training a $\text{Faster R-CNN}_{\textit{n}}$ detector on such fake/synthetic nighttime images with the same annotations of daytime images. Then we test the trained model on the nighttime images for vehicle detection. The comparison results are shown in Table~\ref{tab:eval_nighttime}. We compare the detection results in the form of mAP for each subset of nighttime traffic images and the mean mAP for all of them. Day-to-night image translation methods UNIT \cite{liu2017unsupervised}, CycleGAN \cite{zhu2017unpaired}, and GcGAN \cite{fu2019geometry} perform better than or comparable to the baseline Faster R-CNN which directly tests on nighttime images with daytime model of Sec. \ref{sec:daytime}.
Taking the dataset \textbf{Night4} as an example, the proposed method, based on night-to-day image translation without retraining one more model, achieves the highest 92.94\% mAP, about 5.4\% higher than $\text{Faster R-CNN}_{\textit{n}}$ + $\text{GcGAN}_{\textit{d2n}}$, 3.3\% higher than $\text{Faster R-CNN}_{\textit{n}}$ + $\text{CycleGAN}_{\textit{d2n}}$, 4.8\% higher than $\text{Faster R-CNN}_{\textit{\textit{n}}}$ + $\text{UNIT}_{\textit{d2n}}$ and 5.9\% higher than the baseline Faster R-CNN. The proposed method achieves the best mean mAP with 87.80\% for all nighttime traffic images, despite that $\text{Faster R-CNN}_{\textit{n}}$ + $\text{GcGAN}_{\textit{d2n}}$ performs a little bit better on the \textbf{Night1} subset.
We also provide a traditional method Mean-BGS \cite{li2013video} performing vehicle detection through background subtraction and the daytime model of SSD \cite{liu2016ssd} performing vehicle detection directly on nighttime images. Both of them are worse than Faster R-CNN for nighttime vehicle detection. 
As the corresponding  detection results are shown in Fig.~\ref{fig:vis_n}, we can clearly see that the proposed method is robust to various light conditions. UNIT, CycleGAN and GcGAN based methods could not well detect vehicles under poor light conditions and missed many black vehicles compared to the proposed method, and Faster R-CNN without any image translation does not perform well due to the domain shift of daytime and nighttime scenarios.

\begin{table}[htbp]
\centering
\caption{Nighttime vehicle detection results based on day-to-night translation.
Note that the $\text{Faster R-CNN}_{\textit{n}}$ model is trained on the fake/synthetic nighttime images.}
\resizebox{1\linewidth}{!}{
\begin{tabular}{l|cccc|c}
\toprule
Method ~~$\backslash$~~ \text{mAP(\%)} & \textbf{Night1} & \textbf{Night2} & \textbf{Night3} & \textbf{Night4} & \textbf{Mean} \\
\midrule
\text{Mean-BGS \cite{li2013video}}      
& 54.03 & 49.09 & 52.16 & 55.56 & 52.71 \\
\text{SSD \cite{liu2016ssd}}   & 74.06 & 73.78 & 84.02 & 87.00 & 79.71 \\
\text{Faster R-CNN \cite{ren2015faster}}        
& 74.84 & 74.05 & 85.63 & 87.05 & 80.39 \\
$\text{Faster R-CNN}_{\textit{n}}$ \cite{ren2015faster}+ $\text{UNIT}_{\textit{d2n}}$ \cite{liu2017unsupervised}      
& 70.56 & 77.13 & 82.87 & 88.19 & 79.68 \\ 
$\text{Faster R-CNN}_{\textit{n}}$ \cite{ren2015faster}+ $\text{CycleGAN}_{\textit{d2n}}$ \cite{zhu2017unpaired}  
& 79.39 & 80.72 & 88.72 & 89.66 & 84.62 \\
$\text{Faster R-CNN}_{\textit{n}}$ \cite{ren2015faster}+ $\text{GcGAN}_{\textit{d2n}}$ \cite{fu2019geometry}  
& \textbf{80.89} & 84.20  & 83.92 & 87.55 & 84.14 \\
\textbf{Proposed}   
& 80.25 & \textbf{84.81} & \textbf{93.20} & \textbf{92.94} & \textbf{87.80} \\
\bottomrule
\end{tabular} 
}
\label{tab:eval_nighttime}
\end{table}

\begin{figure*}[th!]
\centering
\includegraphics[width=1.0\textwidth]{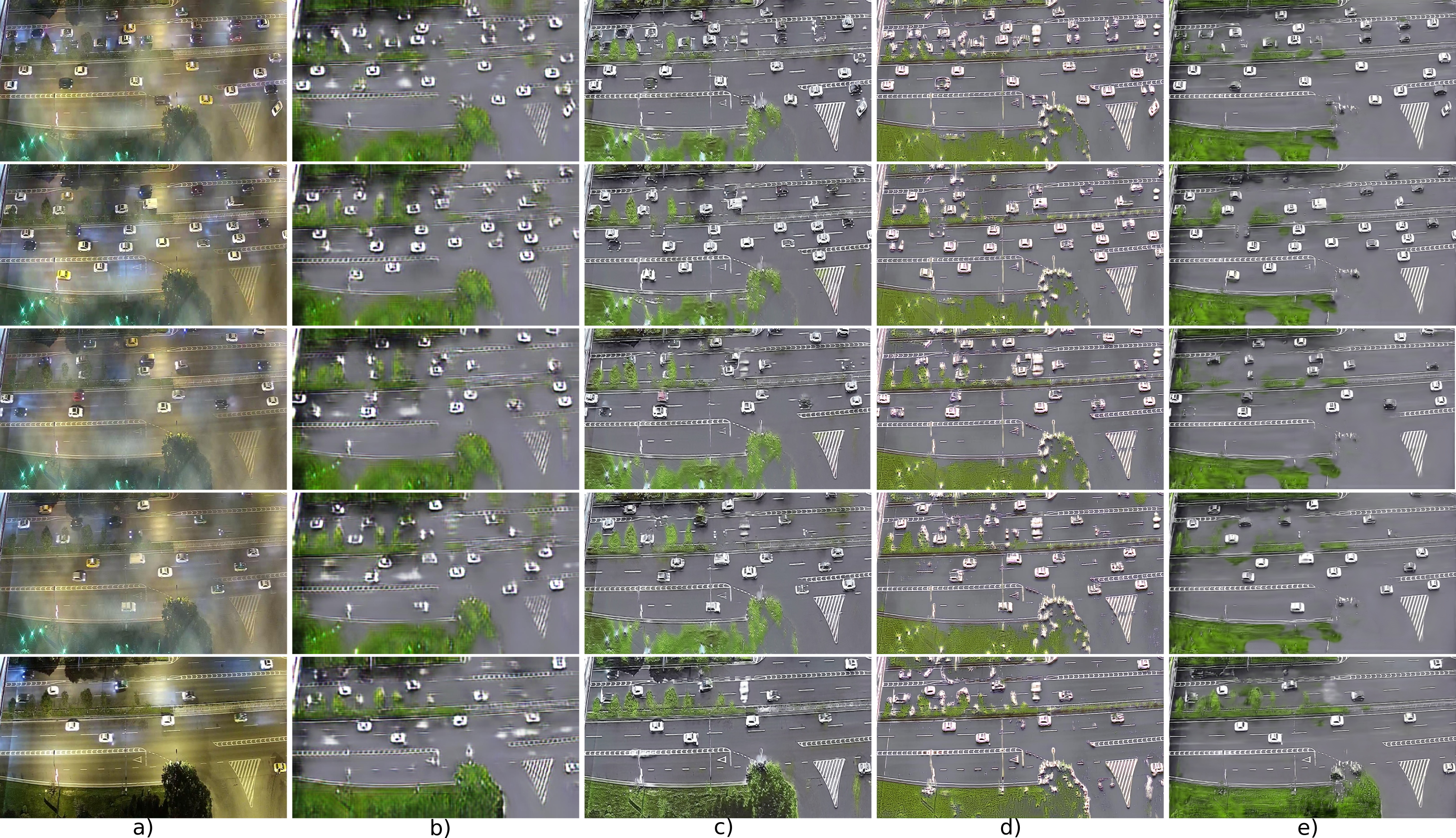}
\caption{Visualization results of image translation from nighttime to daytime. a) is target nighttime image, b) to e) are the image translation results of $\text{UNIT}_{n2d}$ \cite{liu2017unsupervised}, $\text{CycleGAN}_{n2d}$ \cite{zhu2017unpaired}, $\text{GcGAN}_{n2d}$ \cite{fu2019geometry}, and proposed method, respectively.}
\label{fig:vis_d}
\end{figure*}

\subsection{Results compared to night-to-day translation methods}
We also compare our method with these image translation methods in a night-do-day direction. Comparison translation methods, $\text{UNIT}_{\textit{n2d}}$~\cite{liu2017unsupervised}, $\text{CycleGAN}_{\textit{n2d}}$~\cite{zhu2017unpaired}, and $\text{GcGAN}_{\textit{n2d}}$~\cite{fu2019geometry}, first translate nighttime images to daytime-style images, and then these daytime-style images are fed into the daytime model, obtained from Sec. \ref{sec:daytime}, for vehicle detection. The results are shown in Table~\ref{tab:eval_nighttime_n2d}. It shows that the proposed method could achieve the best mean mAP performance than UNIT, CycleGAN, and GcGAN for nighttime vehicle detection via night-to-day translation, demonstrating the advances of the proposed method. This is because the proposed method considers the per-pixel kernel fusion of neighboring information for each pixel and the object detection task during image translation training, preserving more features, \eg, structure details, which are critical for the detection task.

\begin{table}[htbp]
\centering
\caption{Nighttime vehicle detection based on night-to-day translation.}\label{tab:eval_nighttime_n2d}
\footnotesize
\resizebox{1\linewidth}{!}{
\begin{tabular}{l|cccc|c}
\toprule
Method ~~$\backslash$~~ \text{mAP(\%)} & \textbf{Night1}  & \textbf{Night2} & \textbf{Night3} & \textbf{Night4} & \textbf{Mean} \\
\midrule
\text{Mean-BGS \cite{li2013video}}        
& 54.03 & 49.09 & 52.16 & 55.56 & 52.71 \\
\text{SSD \cite{liu2016ssd}}       
& 74.06 & 73.78 & 84.02 & 87.00 & 79.71 \\
\text{Faster R-CNN \cite{ren2015faster}}        
& 74.84 & 74.05 & 85.63 & 87.05 & 80.39 \\
Faster R-CNN \cite{ren2015faster}+ $\text{UNIT}_{\textit{n2d}}$~\cite{liu2017unsupervised}  
& 62.54 & 62.86 & 81.29 & 81.00 & 71.92\\
Faster R-CNN \cite{ren2015faster}+ $\text{CycleGAN}_{\textit{n2d}}$~\cite{zhu2017unpaired} 
& 77.69 & 79.12 & 88.31 & 88.79  & 83.47\\
Faster R-CNN \cite{ren2015faster}+ $\text{GcGAN}_{\textit{n2d}}$~\cite{fu2019geometry} 
& \textbf{83.57} & 83.80 & 79.26 & 83.94 & 82.64 \\
\textbf{Proposed} 
& 80.25 & \textbf{84.81} & \textbf{93.20} & \textbf{92.94} & \textbf{87.80}  \\
\bottomrule
\end{tabular} 
}
\end{table}

Visualization results by the above mentioned image translation methods from nighttime to daytime are shown in Fig.~\ref{fig:vis_d}. It shows that the proposed method could recover the daytime scenario with details. UNIT method suffers from model collapse presenting poor local textures and details. Specifically, the translated images are blurred, especially for the shape and edge of vehicles inside the image. It is consist with the lower detection performance in the form of mAP in Table \ref{tab:eval_nighttime_n2d}. CycleGAN could translate the texture of the vehicles from nighttime to daytime, but it is not robust to the intense road mirror reflections. It presents fake vehicles which do not exist in source nighttime images, resulting in more false positive detection samples. More black vehicles disappeared in the translated images by CycleGAN. GcGAN is also sensitive to such intense road mirror reflections, resulting in more fake vehicles in the translated images. Although the translated trees from the proposed method suffer from corruption, the goal of our work is to accurately detect vehicles at nighttime, we do not care much more about the tree corruption during the night-to-day translation. It is obvious that the translated cars are more natural with clear structures from the proposed method,  resulting in better detection performance in Table \ref{tab:eval_nighttime_n2d}. This is because the image translation training from nighttime to daytime makes full use of paired synthetic data with pixel-wise correspondence and per-pixel kernel fusion of neighboring information which provides rich spatial context information.

\subsection{Ablation Study}
\begin{table}[htbp]
\centering
\caption{Ablation study of the proposed method for the nighttime vehicle detection.}\label{tab:eval_abl}
\footnotesize
\resizebox{1\linewidth}{!}
{
\begin{tabular}{l|cccccc}
\toprule
Method ~~$\backslash$~~ \text{mAP(\%)} & \textbf{Night1}   
& \textbf{Night2} & \textbf{Night3} & \textbf{Night4}  \tabularnewline
\midrule
$\text{Baseline}$ 
& 62.51 &  60.97 & 81.29  & 78.66 \\
$\text{Baseline + StyleMix}$ 
& 75.90 & 76.20 & 88.20 & 86.90 \\
$\text{Baseline + StyleMix + Zero}$ 
& 79.50 & 77.40 & 89.70 & 89.80 \\
$\text{Baseline + StyleMix + Contrast}$ 
& \textbf{80.20} & \textbf{81.79} & \textbf{92.11} & \textbf{91.58} \tabularnewline
\bottomrule
\end{tabular}
}
\end{table}

In this section, we evaluate the contribution of each step in the proposed method: 1) training KPN without the StyleMix, instead, given each daytime image, we randomly select two style images from five style reference images to generate two synthetic nighttime images based on image translation,   respectively. Two synthetic nighttime images are fed into KPN for training combined with detection task. We view this method as our  \textit{Baseline} in this section. 2) On the basis of step 1, we augment the style references in depth and width, denoted as \textit{Baseline + StyleMix}. There are five style reference images for StyleMix, out of  which two are from the nighttime traffic images of D\&N-Car dataset. 3) For testing phase, we preprocess the target nighttime images by Zero-Reference network \cite{guo2020zero} to improve local contrast, then go through KPN for image translation to daytime scenario and followed by detection task via daytime detection model, denoted as \textit{Baseline + StyleMix + Zero}. 4) Different with the preprocessing of step 4, we enhance the local contrast by improving the pixel value less than a threshold, denoted as \textit{Baseline + StyleMix + Contrast}. The corresponding result of the proposed method in each step on the four night subsets in terms of mAP evaluation metric is shown in Table~\ref{tab:eval_abl}. 

We can clearly see the positive effect of each step with the increasing mAP performance. Taking dataset \textbf{Night1} for an example, the baseline method could achieve 62.51\% in mAP. When augmenting and mixing the style reference images to embody the diversity of synthetic nighttime scenarios, mAP increases by about 13\%. When prepossessing the target nighttime images with Zero-Reference network and Contrast, the mAP continues to increase to 79.50\% and 80.20\%, respectively.

We conduct ablation experiments to verify the effectiveness of the style reference setting for StyleMix. We construct a night-style image pool for style reference selection, which consists of 21 images, 7 from nighttime dataset of D\&N-Car, \ie, nighttime traffic images used in this paper, 7 from BDD dataset \cite{yu2020bdd100k} with nighttime scene and the other 7 from $\text{WCT}^{2}$ publicized project website\footnote{https://github.com/clovaai/WCT2}. We randomly choose 5 images from this night-style image pool as style reference for the whole image translation model. We conduct three experiments with different style reference setting for StyleMix. There are 1, 2, and 5 images from the nighttime images of D\&N-Car dataset as different experiment settings:    $\textit{StyleMix}_{1}$, $\textit{StyleMix}_{2}$ and $\textit{StyleMix}_{5}$. The  detection results are shown in Table \ref{tab:eval_abl_sty}. It turns out that the detection performance increases with more night-style images involved from the D\&N-Car dataset. It is reasonable since we expect the StyleMix model to render the synthetic nighttime image closer to the corresponding nighttime style of nighttime images.

\begin{table}[htbp]
\centering
\caption{Ablation study with different style reference setting.}\label{tab:eval_abl_sty}
\footnotesize
\resizebox{1\linewidth}{!}
{
\begin{tabular}{l|cccc|c}
\toprule
Method ~~$\backslash$~~ \text{mAP(\%)} & \textbf{Night1}   
& \textbf{Night2} & \textbf{Night3} & \textbf{Night4} & \textbf{Mean} \\
\midrule

$\text{Baseline + $\textit{StyleMix}_{1}$+ Contrast}$ 
& 76.27 & 79.51 & 91.76 & 91.69 & 84.80\\
$\text{Baseline + $\textit{StyleMix}_{2}$ + Contrast}$ 
& 80.20 & 81.79 & 92.11 & 91.58 & 86.42\\
$\text{Baseline + $\textit{StyleMix}_{5}$ + Contrast}$ 
& \textbf{80.25} & \textbf{84.81} & \textbf{93.20} &  \textbf{92.94} & \textbf{87.80}\\
\bottomrule
\end{tabular}
}
\end{table}

\section{Conclusions}\label{sec:concl}
In this paper, we proposed a detail-preserving method to implement the nighttime to daytime image translation and thus adapt daytime trained detection model to nighttime detection. We firstly utilize style translation method to acquire
paired images of daytime and nighttime, which are hard to obtain in real-world applications. We propose to stylemix the reference styles to embody the diversity of synthetic nighttime scenarios. The following nighttime to daytime translation is implemented based on kernel prediction network to avoid texture corruption and trained with detection task to make the translated daytime image not only visually  photo-realistic to the daytime scenario but also fit the detection task to reuse the daytime domain knowledge. The proposed method can perform both daytime and nighttime vehicle detection with one model. Experimental results showed that the proposed method achieved effective and accurate nighttime detection results.

\bibliographystyle{IEEEtran}
\bibliography{ref}

\begin{thebibliography}{10}
\providecommand{\url}[1]{#1}
\csname url@samestyle\endcsname
\providecommand{\newblock}{\relax}
\providecommand{\bibinfo}[2]{#2}
\providecommand{\BIBentrySTDinterwordspacing}{\spaceskip=0pt\relax}
\providecommand{\BIBentryALTinterwordstretchfactor}{4}
\providecommand{\BIBentryALTinterwordspacing}{\spaceskip=\fontdimen2\font plus
\BIBentryALTinterwordstretchfactor\fontdimen3\font minus
  \fontdimen4\font\relax}
\providecommand{\BIBforeignlanguage}[2]{{%
\expandafter\ifx\csname l@#1\endcsname\relax
\typeout{** WARNING: IEEEtran.bst: No hyphenation pattern has been}%
\typeout{** loaded for the language `#1'. Using the pattern for}%
\typeout{** the default language instead.}%
\else
\language=\csname l@#1\endcsname
\fi
#2}}
\providecommand{\BIBdecl}{\relax}
\BIBdecl

\bibitem{zhang2020unsupervised}
Y.~Zhang, X.~Liang, D.~Zhang, M.~Tan, and E.~P. Xing, ``Unsupervised
  object-level video summarization with online motion auto-encoder,''
  \emph{Pattern Recognition Letters}, vol. 130, pp. 376--385, 2020.

\bibitem{zhang2018spftn}
D.~Zhang, J.~Han, L.~Yang, and D.~Xu, ``Spftn: a joint learning framework for
  localizing and segmenting objects in weakly labeled videos,'' \emph{IEEE
  Transactions on Pattern Analysis and Machine Intelligence}, 2018.

\bibitem{yang2018segmentation}
L.~Yang, J.~Han, D.~Zhang, N.~Liu, and D.~Zhang, ``Segmentation in weakly
  labeled videos via a semantic ranking and optical warping network,''
  \emph{IEEE Transactions on Image Processing}, vol.~27, no.~8, pp. 4025--4037,
  2018.

\bibitem{zhang2018poseflow}
D.~Zhang, G.~Guo, D.~Huang, and J.~Han, ``Poseflow: A deep motion
  representation for understanding human behaviors in videos,'' in \emph{IEEE
  Conference on Computer Vision and Pattern Recognition}, 2018, pp. 6762--6770.

\bibitem{han2018reinforcement}
J.~Han, L.~Yang, D.~Zhang, X.~Chang, and X.~Liang, ``Reinforcement
  cutting-agent learning for video object segmentation,'' in \emph{IEEE
  Conference on Computer Vision and Pattern Recognition}, 2018, pp. 9080--9089.

\bibitem{yu2020weakly}
H.~Yu, D.~Guo, Z.~Yan, L.~Fu, J.~Simmons, C.~P. Przybyla, and S.~Wang, ``Weakly
  supervised easy-to-hard learning for object detection in image sequences,''
  \emph{Neurocomputing}, vol. 398, pp. 71--82, 2020.

\bibitem{li2014video}
S.~Li, H.~Yu, J.~Zhang, K.~Yang, and R.~Bin, ``Video-based traffic data
  collection system for multiple vehicle types,'' \emph{IET Intelligent
  Transport Systems}, vol.~8, no.~2, pp. 164--174, 2014.

\bibitem{ke2018real}
R.~Ke, Z.~Li, J.~Tang, Z.~Pan, and Y.~Wang, ``Real-time traffic flow parameter
  estimation from uav video based on ensemble classifier and optical flow,''
  \emph{IEEE Transactions on Intelligent Transportation Systems}, vol.~20,
  no.~1, pp. 54--64, 2018.

\bibitem{chen2020high}
X.~Chen, Z.~Li, Y.~Yang, L.~Qi, and R.~Ke, ``High-resolution vehicle trajectory
  extraction and denoising from aerial videos,'' \emph{IEEE Transactions on
  Intelligent Transportation Systems}, 2020.

\bibitem{jinlong2020domain}
J.~Li, Z.~Xu, L.~Fu, X.~Zhou, and H.~Yu, ``Domain adaptation from daytime to
  nighttime: A situation-sensitive vehicle detection and traffic flow parameter
  estimation framework,'' \emph{Transportation Research Part C: Emerging
  Technologies}, vol. 124, p. 102946, 2021.

\bibitem{zhu2017unpaired}
J.-Y. Zhu, T.~Park, P.~Isola, and A.~A. Efros, ``Unpaired image-to-image
  translation using cycle-consistent adversarial networks,'' in \emph{IEEE
  International Conference on Computer Vision}, 2017, pp. 2223--2232.

\bibitem{liu2017unsupervised}
M.-Y. Liu, T.~Breuel, and J.~Kautz, ``Unsupervised image-to-image translation
  networks,'' in \emph{Advances in Neural Information Processing Systems},
  2017, pp. 700--708.

\bibitem{park2020contrastive}
T.~Park, A.~A. Efros, R.~Zhang, and J.-Y. Zhu, ``Contrastive learning for
  unpaired image-to-image translation,'' in \emph{European Conference on
  Computer Vision}.\hskip 1em plus 0.5em minus 0.4em\relax Springer, 2020, pp.
  319--345.

\bibitem{huang2018multimodal}
X.~Huang, M.-Y. Liu, S.~Belongie, and J.~Kautz, ``Multimodal unsupervised
  image-to-image translation,'' in \emph{European Conference on Computer
  Vision}, 2018, pp. 172--189.

\bibitem{wang2019learning}
Q.~Wang, J.~Gao, W.~Lin, and Y.~Yuan, ``Learning from synthetic data for crowd
  counting in the wild,'' in \emph{IEEE Conference on Computer Vision and
  Pattern Recognition}, 2019, pp. 8198--8207.

\bibitem{yao2016semantic}
X.~Yao, J.~Han, G.~Cheng, X.~Qian, and L.~Guo, ``Semantic annotation of
  high-resolution satellite images via weakly supervised learning,'' \emph{IEEE
  Transactions on Geoscience and Remote Sensing}, vol.~54, no.~6, pp.
  3660--3671, 2016.

\bibitem{han2019p}
J.~Han, X.~Yao, G.~Cheng, X.~Feng, and D.~Xu, ``P-cnn: Part-based convolutional
  neural networks for fine-grained visual categorization,'' \emph{IEEE
  Transactions on Pattern Analysis and Machine Intelligence}, 2019.

\bibitem{hendrycks2019augmix}
D.~Hendrycks, N.~Mu, E.~D. Cubuk, B.~Zoph, J.~Gilmer, and B.~Lakshminarayanan,
  ``Augmix: A simple data processing method to improve robustness and
  uncertainty,'' in \emph{International Conference on Learning
  Representations}, 2019.

\bibitem{liu2016ssd}
W.~Liu, D.~Anguelov, D.~Erhan, C.~Szegedy, S.~Reed, C.-Y. Fu, and A.~C. Berg,
  ``Ssd: Single shot multibox detector,'' in \emph{European Conference on
  Computer Vision}.\hskip 1em plus 0.5em minus 0.4em\relax Springer, 2016, pp.
  21--37.

\bibitem{redmon2016you}
J.~Redmon, S.~Divvala, R.~Girshick, and A.~Farhadi, ``You only look once:
  Unified, real-time object detection,'' in \emph{IEEE International Conference
  on Computer Vision}, 2016, pp. 779--788.

\bibitem{lin2017focal}
T.-Y. Lin, P.~Goyal, R.~Girshick, K.~He, and P.~Doll{\'a}r, ``Focal loss for
  dense object detection,'' in \emph{IEEE International Conference on Computer
  Vision}, 2017, pp. 2980--2988.

\bibitem{ren2015faster}
S.~Ren, K.~He, R.~Girshick, and J.~Sun, ``Faster r-cnn: Towards real-time
  object detection with region proposal networks,'' in \emph{Advances in Neural
  Information Processing Systems}, 2015, pp. 91--99.

\bibitem{he2017mask}
K.~He, G.~Gkioxari, P.~Doll{\'a}r, and R.~Girshick, ``Mask r-cnn,'' in
  \emph{IEEE International Conference on Computer Vision}, 2017, pp.
  2961--2969.

\bibitem{xu2005pedestrian}
F.~Xu, X.~Liu, and K.~Fujimura, ``Pedestrian detection and tracking with night
  vision,'' \emph{IEEE Transactions on Intelligent Transportation Systems},
  vol.~6, no.~1, pp. 63--71, 2005.

\bibitem{ge2009real}
J.~Ge, Y.~Luo, and G.~Tei, ``Real-time pedestrian detection and tracking at
  nighttime for driver-assistance systems,'' \emph{IEEE Transactions on
  Intelligent Transportation Systems}, vol.~10, no.~2, pp. 283--298, 2009.

\bibitem{choi2018kaist}
Y.~Choi, N.~Kim, S.~Hwang, K.~Park, J.~S. Yoon, K.~An, and I.~S. Kweon, ``Kaist
  multi-spectral day/night data set for autonomous and assisted driving,''
  \emph{IEEE Transactions on Intelligent Transportation Systems}, vol.~19,
  no.~3, pp. 934--948, 2018.

\bibitem{kuang2017bayes}
H.~Kuang, K.-F. Yang, L.~Chen, Y.-J. Li, L.~L.~H. Chan, and H.~Yan, ``Bayes
  saliency-based object proposal generator for nighttime traffic images,''
  \emph{IEEE Transactions on Intelligent Transportation Systems}, vol.~19,
  no.~3, pp. 814--825, 2017.

\bibitem{satzoda2016looking}
R.~K. Satzoda and M.~M. Trivedi, ``Looking at vehicles in the night: Detection
  and dynamics of rear lights,'' \emph{IEEE Transactions on Intelligent
  Transportation Systems}, 2016.

\bibitem{alvarez2010road}
J.~M.~{\'A}. Alvarez and A.~M. {\'L}opez, ``Road detection based on illuminant
  invariance,'' \emph{IEEE Transactions on Intelligent Transportation Systems},
  vol.~12, no.~1, pp. 184--193, 2010.

\bibitem{ros2015unsupervised}
G.~Ros and J.~M. Alvarez, ``Unsupervised image transformation for outdoor
  semantic labelling,'' in \emph{IEEE Intelligent Vehicles Symposium}.\hskip
  1em plus 0.5em minus 0.4em\relax IEEE, 2015, pp. 537--542.

\bibitem{valada2017adapnet}
A.~Valada, J.~Vertens, A.~Dhall, and W.~Burgard, ``Adapnet: Adaptive semantic
  segmentation in adverse environmental conditions,'' in \emph{IEEE
  International Conference on Robotics and Automation}.\hskip 1em plus 0.5em
  minus 0.4em\relax IEEE, 2017, pp. 4644--4651.

\bibitem{anoosheh2019night}
A.~Anoosheh, T.~Sattler, R.~Timofte, M.~Pollefeys, and L.~Van~Gool,
  ``Night-to-day image translation for retrieval-based localization,'' in
  \emph{IEEE International Conference on Robotics and Automation}.\hskip 1em
  plus 0.5em minus 0.4em\relax IEEE, 2019, pp. 5958--5964.

\bibitem{tian2020bias}
B.~Tian, F.~Juefei-Xu, Q.~Guo$^{*}$, W.~Chan, Y.~Cheng, X.~Li, X.~Xie, and
  S.~Qin, ``Bias field poses a threat to dnn-based x-ray recognition,'' in
  \emph{IEEE International Conference on Multimedia and Expo}, 2021.

\bibitem{neurips20_abba}
Q.~Guo, F.~Juefei-Xu, X.~Xie, L.~Ma, J.~Wang, B.~Yu, W.~Feng, and Y.~Liu,
  ``{Watch out! Motion is Blurring the Vision of Your Deep Neural Networks},''
  in \emph{Advances in Neural Information Processing Systems}, 2020.

\bibitem{guo2021exploring}
Q.~Guo, W.~Feng, R.~Gao, Y.~Liu, and S.~Wang, ``Exploring the effects of blur
  and deblurring to visual object tracking,'' \emph{IEEE Transactions on Image
  Processing}, vol.~30, pp. 1812--1824, 2021.

\bibitem{guo2020selective}
Q.~Guo, R.~Han, W.~Feng, Z.~Chen, and L.~Wan, ``Selective spatial
  regularization by reinforcement learned decision making for object
  tracking,'' \emph{IEEE Transactions on Image Processing}, vol.~29, pp.
  2999--3013, 2020.

\bibitem{guo2020spark}
Q.~Guo, X.~Xie, F.~Juefei-Xu, L.~Ma, Z.~Li, W.~Xue, W.~Feng, and Y.~Liu,
  ``Spark: Spatial-aware online incremental attack against visual tracking,''
  in \emph{European Conference on Computer Vision}, vol.~2, 2020.

\bibitem{guo2017learning}
Q.~Guo, W.~Feng, C.~Zhou, R.~Huang, L.~Wan, and S.~Wang, ``Learning dynamic
  siamese network for visual object tracking,'' in \emph{IEEE International
  Conference on Computer Vision}, 2017, pp. 1763--1771.

\bibitem{zhang2018fully}
Y.~Zhang, Z.~Qiu, T.~Yao, D.~Liu, and T.~Mei, ``Fully convolutional adaptation
  networks for semantic segmentation,'' in \emph{IEEE Conference on Computer
  Vision and Pattern Recognition}, 2018, pp. 6810--6818.

\bibitem{bak2018domain}
S.~Bak, P.~Carr, and J.-F. Lalonde, ``Domain adaptation through synthesis for
  unsupervised person re-identification,'' in \emph{European Conference on
  Computer Vision}, 2018, pp. 189--205.

\bibitem{zhang2020cross}
P.~Zhang, B.~Zhang, D.~Chen, L.~Yuan, and F.~Wen, ``Cross-domain correspondence
  learning for exemplar-based image translation,'' in \emph{IEEE Conference on
  Computer Vision and Pattern Recognition}, 2020, pp. 5143--5153.

\bibitem{kim2020learning}
M.~Kim and H.~Byun, ``Learning texture invariant representation for domain
  adaptation of semantic segmentation,'' in \emph{IEEE Conference on Computer
  Vision and Pattern Recognition}, 2020, pp. 12\,975--12\,984.

\bibitem{zheng2020cross}
Y.~Zheng, D.~Huang, S.~Liu, and Y.~Wang, ``Cross-domain object detection
  through coarse-to-fine feature adaptation,'' in \emph{IEEE Conference on
  Computer Vision and Pattern Recognition}, 2020, pp. 13\,766--13\,775.

\bibitem{anoosheh2018combogan}
A.~Anoosheh, E.~Agustsson, R.~Timofte, and L.~Van~Gool, ``Combogan:
  Unrestrained scalability for image domain translation,'' in \emph{IEEE
  Conference on Computer Vision and Pattern Recognition Workshops}, 2018, pp.
  783--790.

\bibitem{romero2019smit}
A.~Romero, P.~Arbel{\'a}ez, L.~Van~Gool, and R.~Timofte, ``Smit: Stochastic
  multi-label image-to-image translation,'' in \emph{IEEE International
  Conference on Computer Vision Workshops}, 2019, pp. 0--0.

\bibitem{fu2019geometry}
H.~Fu, M.~Gong, C.~Wang, K.~Batmanghelich, K.~Zhang, and D.~Tao,
  ``Geometry-consistent generative adversarial networks for one-sided
  unsupervised domain mapping,'' in \emph{IEEE Conference on Computer Vision
  and Pattern Recognition}, 2019, pp. 2427--2436.

\bibitem{cai2019toward}
J.~Cai, H.~Zeng, H.~Yong, Z.~Cao, and L.~Zhang, ``Toward real-world single
  image super-resolution: A new benchmark and a new model,'' in \emph{IEEE
  International Conference on Computer Vision}, 2019, pp. 3086--3095.

\bibitem{mildenhall2018burst}
B.~Mildenhall, J.~T. Barron, J.~Chen, D.~Sharlet, R.~Ng, and R.~Carroll,
  ``Burst denoising with kernel prediction networks,'' in \emph{IEEE Conference
  on Computer Vision and Pattern Recognition}, 2018, pp. 2502--2510.

\bibitem{guo2020efficientderain}
Q.~Guo, J.~Sun, F.~Juefei-Xu, L.~Ma, X.~Xie, W.~Feng, and Y.~Liu,
  ``Efficientderain: Learning pixel-wise dilation filtering for high-efficiency
  single-image deraining,'' in \emph{AAAI Conference on Artificial
  Intelligence}, 2021.

\bibitem{fu2021auto}
L.~Fu, C.~Zhou, Q.~Guo, F.~Juefei-Xu, H.~Yu, W.~Feng, Y.~Liu, and S.~Wang,
  ``Auto-exposure fusion for single-image shadow removal,'' in \emph{IEEE
  Conference on Computer Vision and Pattern Recognition}, 2021.

\bibitem{girshick2014rich}
R.~Girshick, J.~Donahue, T.~Darrell, and J.~Malik, ``Rich feature hierarchies
  for accurate object detection and semantic segmentation,'' in \emph{IEEE
  Conference on Computer Vision and Pattern Recognition}, 2014, pp. 580--587.

\bibitem{li2013video}
S.~Li, H.~Yu, J.~Zhang, K.~Yang, and R.~Bin, ``Video-based traffic data
  collection system for multiple vehicle types,'' \emph{IET Intelligent
  Transport Systems}, vol.~8, no.~2, pp. 164--174, 2013.

\bibitem{guo2020zero}
C.~Guo, C.~Li, J.~Guo, C.~C. Loy, J.~Hou, S.~Kwong, and R.~Cong,
  ``Zero-reference deep curve estimation for low-light image enhancement,'' in
  \emph{IEEE Conference on Computer Vision and Pattern Recognition}, 2020, pp.
  1780--1789.

\bibitem{yu2020bdd100k}
F.~Yu, H.~Chen, X.~Wang, W.~Xian, Y.~Chen, F.~Liu, V.~Madhavan, and T.~Darrell,
  ``Bdd100k: A diverse driving dataset for heterogeneous multitask learning,''
  in \emph{IEEE Conference on Computer Vision and Pattern Recognition}, 2020,
  pp. 2636--2645.

\end{thebibliography}

\vspace{-30pt}
\begin{IEEEbiography}[{\includegraphics[width=1in,height=1.25in,clip]{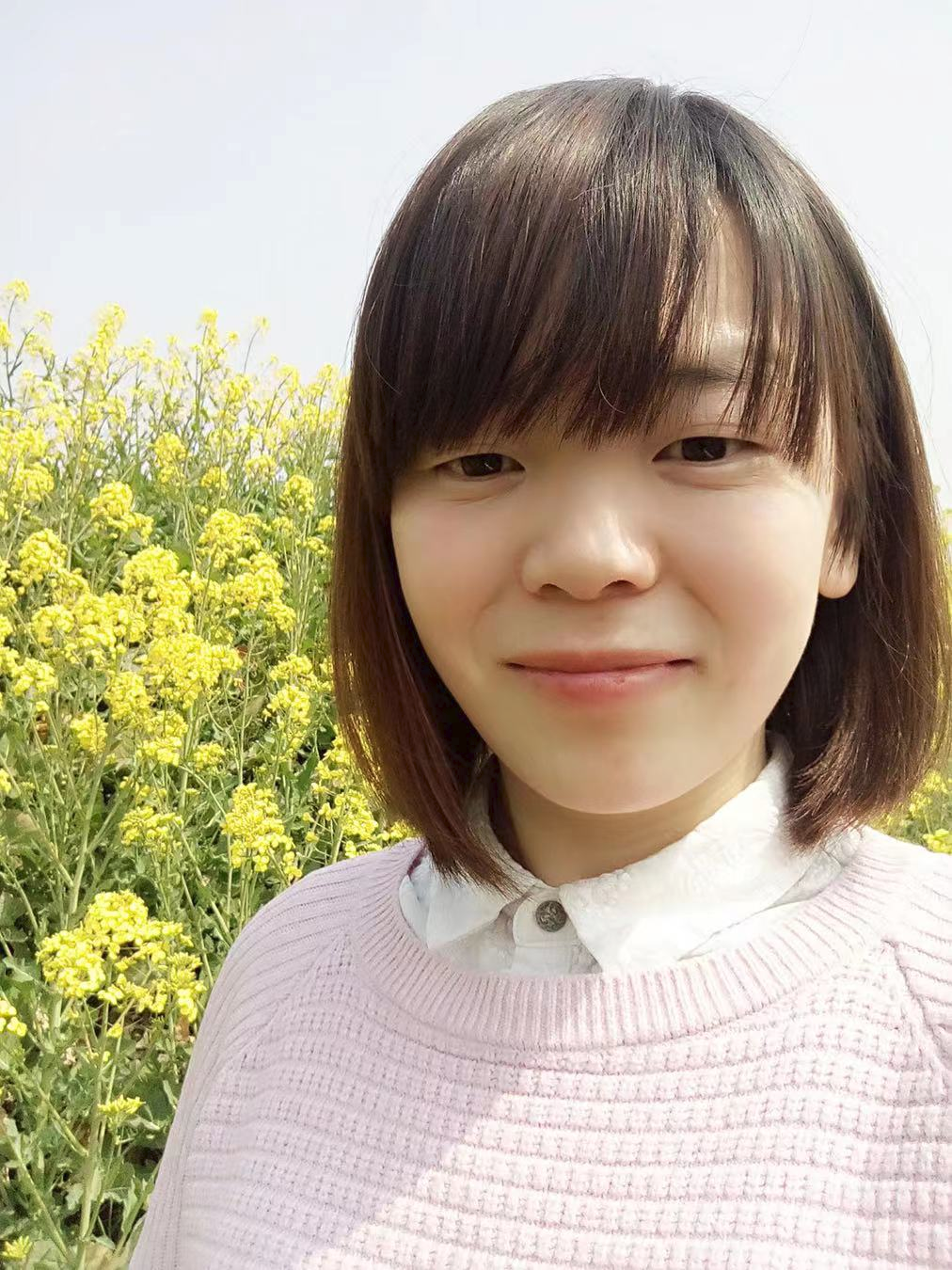}}]{Lan Fu} is pursuing her Ph.D degree in the department of computer science and engineering at University of South Carolina, Columbia, SC, USA. Prior to that, she received the M.S. degree in Biomedical Engineering from Tianjin University, Tianjin, China. Her research interests include computer vision and deep learning and mainly focus on domain adaptation based object detection and image enhancement.
\end{IEEEbiography}   

\begin{IEEEbiography}[{\includegraphics[width=1in,height=1.25in,clip]{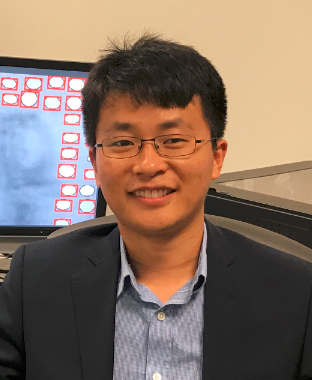}}]{Hongkai Yu} received the Ph.D. degree in computer science and engineering from University of South Carolina, Columbia, SC, USA, in 2018. He is currently an Assistant Professor in the Department of Electrical Engineering and Computer Science at Cleveland State University, Cleveland, OH, USA. His research interests include computer vision, machine learning, deep learning and intelligent transportation system.  
\end{IEEEbiography}    

\begin{IEEEbiography}[{\includegraphics[width=1in,height=1.25in,clip]{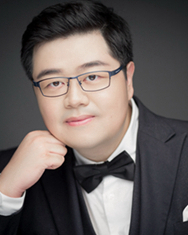}}]{Felix Juefei-Xu} received the Ph.D. degree in Electrical and Computer Engineering from Carnegie Mellon University (CMU), Pittsburgh, PA, USA. Prior to that, he received the M.S. degree in Electrical and Computer Engineering and the M.S degree in Machine Learning from CMU, and the B.S. degree in Electronic Engineering from Shanghai Jiao Tong University (SJTU), Shanghai, China. Currently, he is a Research Scientist with the Alibaba Group, Sunnyvale, CA, USA, with research focus on a fuller understanding of deep learning where he is actively exploring new methods in deep learning that are statistically efficient and adversarially robust. He also has broader interests in pattern recognition, computer vision, machine learning, optimization, statistics, compressive sensing, and image processing. He is the recipient of multiple best/distinguished paper awards, including IJCB'11, BTAS'15-16, ASE'18, and ACCV'18.
\end{IEEEbiography}

\begin{IEEEbiography}[{\includegraphics[width=1in,height=1.25in,clip]{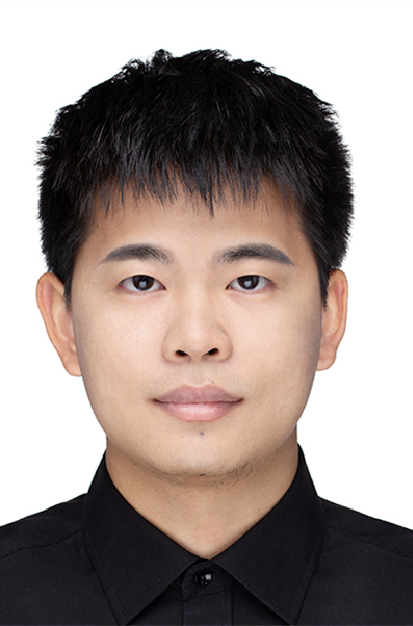}}]{Jinlong Li} received the B.S. degree and M.S. degrees from Chang'an University, Xi'an, China in 2018 and 2021 respectively. His research interests include intelligent transportation system, computer vision, and deep learning. 
\end{IEEEbiography}

\begin{IEEEbiography}[{\includegraphics[width=1in,height=1.25in,clip,keepaspectratio]{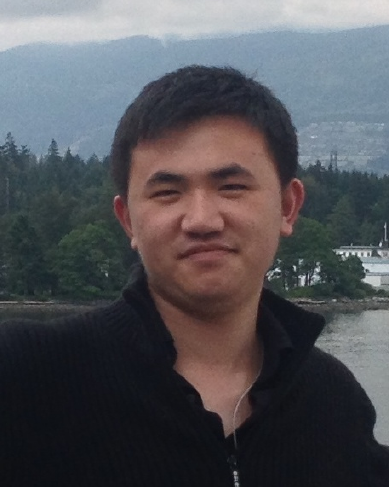}}]{Qing Guo} received his B.S. degree in Electronic and Information Engineering from the North China Institute of Aerospace Engineering in 2011, M.E. degree in computer application technology from the College of Computer and Information Technology, China Three Gorges University in 2014, and the Ph.D. degree in computer application technology from the School of Computer Science and Technology, Tianjin University, China. He was a research fellow with the Nanyang Technology University, Singapore, from Dec. 2019 to Sep. 2020. He is currently a Wallenberg-NTU Presidential Postdoctoral Fellow with the Nanyang Technological University, Singapore. His research interests include computer vision, AI security, and image processing. He is a member of IEEE.
\end{IEEEbiography}

\begin{IEEEbiography}[{\includegraphics[width=1in,height=1.25in,clip,keepaspectratio]{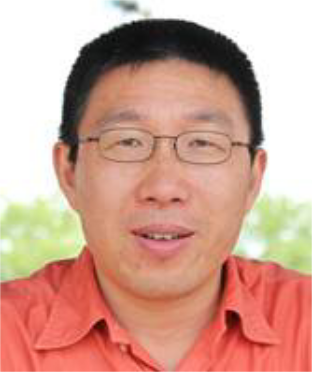}}]{Song Wang} received the Ph.D. degree in electrical and computer engineering from the University of Illinois at Urbana-Champaign (UIUC), Champaign, IL, USA, in 2002. He was a Research Assistant with the Image Formation and Processing Group, Beckman Institute, UIUC, from 1998 to 2002. In 2002, he joined the Department of Computer Science and Engineering, University of South Carolina, Columbia, SC, USA, where he is currently a Professor. His current research interests include computer vision, image processing, and machine learning. Dr. Wang is currently serving as the Publicity/Web Portal Chair of the Technical Committee of Pattern Analysis and Machine Intelligence of the IEEE Computer Society, an Associate Editor of IEEE Transaction on Pattern Analysis and Machine Intelligence, Pattern Recognition Letters, and Electronics Letters. He is a Senior Member of the IEEE and a member of the IEEE Computer Society. \end{IEEEbiography}

\end{document}